\documentclass[runningheads]{llncs}
\usepackage{graphicx}
\usepackage{cite}
\usepackage{amsmath,amssymb,amsfonts}
\usepackage{algorithm,algpseudocode}
\usepackage{algorithmicx}
\algdef{SE}[DOWHILE]{Do}{doWhile}{\algorithmicdo}[1]{\algorithmicwhile\ #1}%
\usepackage{graphicx}
\usepackage{textcomp}
\usepackage{xcolor}
\usepackage{url}
\def\BibTeX{{\rm B\kern-.05em{\sc i\kern-.025em b}\kern-.08em
    T\kern-.1667em\lower.7ex\hbox{E}\kern-.125emX}}

\newcommand{\vect}[1]{\mbox{\boldmath$#1$}}

\def\Amat{\vect{A}}
\def\Bmat{\vect{B}}
\def\Fmat{\vect{F}}
\def\Gmat{\vect{G}}
\def\Xmat{\vect{X}}
\def\Ymat{\vect{Y}}
\def\Zmat{\vect{Z}}
\def\Wmat{\vect{W}}
\def\Smat{\vect{S}}
\def\Emat{\vect{E}}
\def\Imat{\vect{I}}

\def\svec{\vect{s}}
\def\xvec{\vect{x}}
\def\yvec{\vect{y}}
\def\wvec{\vect{w}}
\def\bvec{\vect{b}}

\begin{document}
\title{Efficient Estimation of Unique Components in Independent Component Analysis by Matrix Representation}
\titlerunning{Efficient Estimation of Unique Components in ICA}
%

\author{Yoshitatsu Matsuda\inst{1}\orcidID{0000-0002-0056-0185} \and
Kazunori Yamaguchi\inst{2}\orcidID{0000-0003-0383-7386}}

\authorrunning{Y. Matsuda et al.}
%
\institute{Seikei University, Tokyo, Japan 
\email{matsuda@st.seikei.ac.jp}\\
\and
The University of Tokyo, Tokyo, Japan\\
\email{yamaguch@graco.c.u-tokyo.ac.jp}}
\maketitle              
\begin{abstract}
Independent component analysis (ICA) is a widely used method in various applications of signal processing and feature extraction. It extends principal component analysis (PCA) and can extract important and complicated components with small variances. One of the major problems of ICA is that, unlike PCA, the uniqueness of the solution is not guaranteed. That is because there are many local optima in optimizing the objective function of ICA.
It has been shown previously that the unique global optimum of ICA can be estimated from many random initializations by handcrafted thread computation. In this paper, the unique estimation of ICA is greatly accelerated by reformulating the algorithm in matrix representation and reducing redundant calculations. Experimental results on artificial datasets and EEG data verified the efficiency of the proposed method.
\end{abstract}
\section{Introduction}
Independent component analysis (ICA) is widely used in various applications of signal processing, such as blind source separation \cite{Cichocki.2002,Comon.2010} and feature extraction, such as image decomposition \cite{Hyvarinen.2001}.
The linear model of ICA is given as
$\xvec = \Amat\svec$,
where $\xvec = \left(x_{i}\right)$ is the $N$-dimensional observed
signal, $\Amat = \left(a_{ij}\right)$ is the $N \times N$ mixing matrix, and
$\svec = \left(s_{i}\right)$ is  
the $N$-dimensional source signal.
ICA can estimate $\svec$ from only $\xvec$ without knowing $\Amat$ under the assumption that
the source components are statistically independent of each other
and they are given according to
non-Gaussian distributions.
Under this assumption, it can be shown that the maximization of the non-Gaussianity of each component leads to the estimation of the accurate source.
Higher-order statistics of the components, such as kurtosis, can measure non-Gaussianity.
ICA can be regarded as an extension of principal component analysis (PCA), which maximizes the variances of the components. 
There are various expansions of the simple linear model of ICA (for example, \cite{JMLR.Bach.2002} and \cite{NC.Hao.2010}).
Especially, various flexible models of ICA have recently received attention \cite{IEEEIM.Chen.2017,NI.Hsu.2018,JNE.Banville.2021,JNM.WU.2022}.
However, the linear model described above is still helpful in the practical application, such as the ElectroEncephaloGraphy (EEG) analysis \cite{JNM.Delorme.2004,NBR.Onton.2006}. 
On the other hand, it has been known that the kurtosis-based ICA is quite sensitive to outliers because kurtosis amplifies the extreme values to the fourth power \cite{IEEENN.Hyvarinen.1999}.
Nevertheless, the kurtosis-based ICA has some theoretical and practical superiority \cite{IEEENN.Matsuda.2018,PL1.Matsuda.2022}.
Therefore, this paper focuses only on the linear ICA model based on kurtosis. 

As the objective function of ICA is much more complicated than that of PCA, one of the major problems of ICA is that it is not guaranteed to find the true (globally optimal) solution, unlike PCA. Generally, there are many local optima in the estimation of ICA. To solve this problem, the ordering ICA has been proposed\cite{IEEENN.Matsuda.2018}. The ordering ICA employs a relatively simple objective function of ICA, and it can find the global optimum by a sufficiently large number of threads from random initializations.
The practical usefulness of the ordering ICA has been verified in the application of the EEG analysis
\cite{PL1.Matsuda.2022}.
However, it was not as fast as the simple ICA and there was still significant room for acceleration. In addition, a hand-crafted thread implementation is required for each computing environment. This problem can hinder the availability of the ordering ICA.

In this paper, we propose a novel fast algorithm for the ordering ICA by reformulating it into matrix representations. Matrix manipulations (for example, the multiplication of large matrices) can be calculated efficiently in many software environments.  
Therefore, the proposed algorithm enables the ordering ICA to be easily utilized in various situations.
Though this idea of using matrix manipulation is simple, it is expected to drastically improve the performance of the ordering ICA, similar to the employment of mini-batch optimization in deep learning. 

This paper is organized as follows.  Section \ref{background} briefly describes ICA and the ordering ICA.
In Section \ref{algorithm}, a new algorithm for the ordering ICA is constructed by matrix representation. Section \ref{results} shows the experimental results of the proposed algorithm compared to the other methods using artificial datasets and EEG data.  Lastly, we conclude this paper in Section \ref{conclusion}.

\section{Background}\label{background}
 \subsection{Independent Component Analysis}
Generally, ICA is a method for extracting unknown sources from only the observed signals without knowing the mixture system \cite{Cichocki.2002,Comon.2010}.
In many cases, ICA assumes the linear mixture model
$\Xmat = \Amat\Smat$,
where an $N \times M$ matrix $\Xmat =\left(x_{im}\right)$ 
is the observed signals of given samples
($i = 1, \cdots, N$ and $m = 1, \cdots, M$ denote the indices of components and samples, respectively).
$\Amat = \left(a_{ij}\right)$ (an $N \times N$ matrix) is the unknown mixing matrix.
$\Smat = \left(s_{im}\right)$ (an $N \times M$ matrix) consists of the unknown sources.
The estimation of the source $\Ymat = \left(y_{im}\right)$ is given as $\Ymat = \Wmat\Xmat$, where $\Wmat = \left(w_{ij}\right)$ (an $N \times N$ matrix) is the separating matrix. $\Wmat = \Amat^{-1}$ holds in the true estimation.
The objective functions of ICA $\Phi\left(\Ymat\right)$ are constructed by some higher-order statistics of each component of $\Ymat$ so that
the optimization of $\Phi$ with respect to $\Wmat$ corresponds to the maximization of statistical independence among the components. One of the simplest examples of the objective functions is given as $\Phi = \sum_{i}\sum_{m}y_{im}^{4}$. 

Though many methods of ICA have been proposed depending on the choice of the objective function and the optimization algorithm, there exist three representative methods:
the extended InfoMax \cite{NC.Lee.1999} (referred to as InfoMax),
Fast ICA \cite{IEEENN.Hyvarinen.1999},
and JADE \cite{IEEPF.Cardoso.1993}. 
The majority of other methods are variations of these three methods.
It is guaranteed in these methods that the true estimation of sources is included in the solutions.
However, these methods find many solutions due to local optima.
Therefore, it is not guaranteed that the actual solution is the unique true estimation. 
 
 \subsection{Ordering ICA}
The ordering ICA has been recently proposed for estimating the unique solution in ICA
\cite{IEEENN.Matsuda.2018}. The following objective function is essential in the ordering ICA:
\begin{equation}
     \Upsilon\left(\alpha_{i}\right)=
     \alpha_{i} - 2\log\left(\alpha_{i}/2 + 1\right)
    \label{eq_back_Upsilon_1}
\end{equation}
where 
\begin{equation}
\alpha_{i} = \sum_{m}y_{im}^{4}/M - 3.
\label{eq_back_alpha_1}
\end{equation}
$\alpha_{i}$ is the estimated kurtosis of the $i$-th component of $\Ymat$ if the component is normalized.
When $\Xmat$ is whitened in advance and $\Wmat$ is constrained to orthonormal, the following theorem holds:
 \begin{theorem}\label{theorem_1}
 Under the assumption that $\Smat$ consists of any normalized independent sources (other than the uniform Bernoulli variable) and $\Xmat$ is given by an invertible linear ICA model
 $\Xmat = \Amat\Smat$, 
 all the non-Gaussian sources are extracted 
 by maximizing globally $\Upsilon\left(\alpha_{i} = \sum_{m}y_{im}^{4}/M - 3\right)$
 for each $i$ subject to $\sum_{m}y_{im} = 0$ and
 $\sum_{m}y_{im}y_{jm}/M = \delta_{ij}$ for $j \le i$.
 In addition, the extracted sources are arranged
 in descending order
 of $\Upsilon\left(\kappa_{i}\right)$, where $\kappa_{i}$ is the true kurtosis of the $i$-th component
 of $\Smat$.
\end{theorem}
 This theorem means that the unique true estimation can be achieved by maximizing globally $\Upsilon\left(\alpha_{i}\right)$ one by one for $i = 1, \cdots, N$ with respect to $\wvec_{i} = \left(w_{i1},\ldots,w_{iN}\right)$.
Though $\Upsilon\left(\alpha_{i}\right)$ has many local maxima, the global maximum can be searched from many different (for example, random) initializations by parallel computing. In addition, each local maximum can be calculated efficiently by the usual Fast ICA algorithm in the deflation approach \cite{IEEENN.Hyvarinen.1999}, which optimizes $\sum_{m}y_{im}^4$ by the Newton-Raphson method one by one under the orthonormality constraints. 
In addition, the ordering ICA can estimate the number of non-Gaussian sources by the Gaussianity test \cite{IEEENN.Matsuda.2018}.
It can be shown that the rest of the components are Gaussian once the following condition is satisfied:
\begin{equation}
\Upsilon\left(\alpha_{i}\right) < \frac{2\left(N-i+2\right)\left(N-i+1\right)}{M}.
\end{equation}

In summary, the original algorithm for the ordering ICA is given in Algorithm \ref{alg_background_original} \cite{IEEENN.Matsuda.2018,PL1.Matsuda.2022}. Here, the observed signal $\Xmat$ is whitened in advance. The hyperparameter $L$ is the number of random initializations. $K$ (the maximum of iterations of the Newton-Raphson method) and $\epsilon$ (the convergence threshold) are the hyperparameters on the termination decision of each run of Fast ICA. $\yvec \circ \yvec$ denotes the element-wise product. Because $L$ runs can be executed simultaneously by parallel computing, this algorithm is as efficient as Fast ICA when the computational resources provide a sufficient number of threads. In the experiments of EEG analysis, this algorithm could find the unique solution approximately if $L$ is about one hundred \cite{PL1.Matsuda.2022}. 
 \begin{algorithm} 
  \caption{Original Ordering ICA with Gaussianity Test (cited from \cite{PL1.Matsuda.2022} with some modifications)}\label{alg_background_original}
    \begin{algorithmic}[1]
    \Require{$\Xmat$: whitened observed signals. $L, K, \epsilon$: hyperparameters.}
    \Ensure{$\Wmat$: separating matrix.}
     \State Initialize $\Wmat$ to the empty matrix
     \State $i \gets 1$
     \While{$i \le N$}
     \If{$i > 1$}
     \State $\Emat \gets \Wmat^{\intercal}\Wmat$
     \Else
     \State $\Emat \gets 0$
     \EndIf
       \State Set $L$ random initializations 
       $\wvec_{i}^{l}$
     \ForAll{$l \in \{1, \ldots, L\}$} (by parallel computing of $L$ runs)
     \State $\hat{\wvec}_{i}^{l}$, $\yvec$ $\leftarrow$\Call{FastICA}{$\wvec_{i}^{l}$, $\Xmat$, $\Emat$}
     \State $\alpha_{il} \gets \sum_{m}y_{m}^{4}/M - 3$
   \EndFor
    \State $p \gets \mathrm{argmax}_{l}\Upsilon\left(\alpha_{il}\right)$
    \If{$\Upsilon\left(\alpha_{ip}\right) < 2\left(N-i+2\right)\left(N-i+1\right)/M$}
    \State \textbf{break}
    \EndIf
     \State Concatenate $\left(\hat{\wvec}_{i}^{p}\right)^{\intercal}$ to $\Wmat$ as the last row
     \State $i \gets i + 1$
     \EndWhile
     \Function{FastICA}{$\wvec$, $\Xmat$, $\Emat$}
     \State $\wvec \gets \wvec - \Emat\wvec$
     \State $\wvec \gets \wvec/\sqrt{\wvec^{\intercal}\wvec}$
     \State $t \gets 0$
     \Do
     \State $\wvec_{\rm{prev}} \gets \wvec$
     \State $\yvec \gets \wvec^{\intercal}\Xmat$
     \State $\wvec \gets \Xmat\left(\yvec \circ \yvec
     \circ \yvec\right)^{\intercal}/M - 3\wvec$
     \State $\wvec \gets \wvec - \Emat\wvec$
     \State $\wvec \gets \wvec/\sqrt{\wvec^{\intercal}\wvec}$
     \State $t \gets t + 1$
     \doWhile{$t < K$ and ($||\wvec + \wvec_{\rm{prev}}|| > \epsilon$ or
       $||\wvec - \wvec_{\rm{prev}}|| > \epsilon$)}
     \State $\yvec \gets \wvec^{\intercal}\Xmat$
     \State\Return{$\wvec$, $\yvec$}
    \EndFunction
   \end{algorithmic}
 \end{algorithm}
 
\section{Algorithm}\label{algorithm}
Here, we describe a novel algorithm for the ordering ICA using matrix representation.
It has three improvements over Algorithm \ref{alg_background_original}.
First, to utilize the efficiency of matrix multiplication, we construct a matrix of partially recovered signals of $L$ runs, as detailed in Section \ref{algorithm_matrix}.
This calculation is almost the same as the symmetrical approach of the usual fast ICA \cite{IEEENN.Hyvarinen.1999} without the orthogonalization process. Though the symmetrical approach is much faster than the usual deflation approach, it is not used in practice due to poor results in the source estimation. Our algorithm revives the symmetrical approach in a different context. 
Second, to reduce the inefficiency caused by keeping the already recovered signals in the matrix, we introduce a dynamic reduction of rows from the matrix, as detailed in Section \ref{algorithm_removal}. 
Third, the orthonormality constraints are enforced by the reduction of $\Xmat$ instead of the Gram-Schmidt orthogonalization of $\wvec$.
It can reduce computational costs according to the number of remaining components (unestimated), as detailed in Section \ref{algorithm_reduction}.

\subsection{Matrix Representation of Multiple Runs}\label{algorithm_matrix}
The most time-consuming part of the ordering ICA is the $L$ runs of the multiplication of $\Xmat$ (an $N \times M$ matrix and $\left(\yvec \circ \yvec \circ \yvec\right)^{\intercal}$ (an $M$-dimensional vector)) in line 28 of Algorithm \ref{alg_background_original}. Let $\Zmat = \left(z_{ij}\right)$ be the $L \times M$ matrix of a set of $\yvec$'s to utilize the matrix manipulation. Similarly, let $\Bmat = \left(b_{ij}\right)$ be the $L \times N$ matrix of $\left(\wvec_{i}^{l}\right)^{\intercal}$ for $l = 1, \cdots, L$.
Thus, the process in lines 27-28 over multiple runs can be executed as a batch by
\begin{equation}
\Zmat \gets \Bmat\Xmat
\end{equation}
and
\begin{equation}
\Bmat \gets \Xmat\left(\Zmat \circ \Zmat
     \circ \Zmat\right)^{\intercal}/M - 3\Bmat.
\end{equation}
Matrix manipulation tools can efficiently calculate them. 

\subsection{Removal of Converged Runs}\label{algorithm_removal}
When a row of $\Bmat$ converges to a local optimum, it is not necessary to update the row further.
In parallel computing, the threads that have finished calculations often have to wait for other threads to finish.  On the other hand, it is enough to reduce the size of $\Bmat$ in matrix representation. $\Bmat$ is separated into only the rows needing more updates (denoted by $\Bmat_{\rm{update}}$) and the other rows corresponding to the converged runs (denoted by $\Bmat_{\rm{conv}}$). $\Bmat_{\rm{conv}}$ can be stored without any updates. Computational resources are concentrated on mandatory calculation, avoiding useless updates of converged rows.

\subsection{Reduction of Signals under Orthonormality Constraints}\label{algorithm_reduction}
In Algorithm \ref{alg_background_original}, the observed signal $\Xmat$ is fixed in a given whitened form. To satisfy the orthonormality constraints, $\wvec$ is orthonormalized in each iteration. On the other hand, the orthogonality can be satisfied by reducing $\Xmat$ to be orthogonal to the already estimated components by an $(i-1) \times N$ matrix $\Wmat$. For this purpose, an $N \times N$ matrix $\Fmat$ is calculated by
 \begin{equation}
 \Fmat = \Imat_{N}-\Wmat^{\intercal}\Wmat,
 \end{equation}
 where $\Imat_{N}$ is the $N$-dimensional identity matrix.
 $\Wmat\Fmat = \vect{0}$ holds because each row of $\Wmat$ is orthonormal to each other. 
 Let $\tilde{\Fmat}$ be an $\left(N - i + 1\right) \times N$ submatrix that includes only the first to the $\left(N - i + 1\right)$-th rows of $\Fmat$. The rows of 
 $\tilde{\Fmat}$ are bases of the orthogonal complement space to $\Wmat$.  Let $\Gmat$ be the orthonormalized $\tilde{\Fmat}$, which is calculated by 
 \begin{equation}
 \Gmat = \left(\tilde{\Fmat}\tilde{\Fmat}^{\intercal}\right)^{-\frac{1}{2}}\tilde{\Fmat}.
 \end{equation}
 Then, $\tilde{\Xmat}$ is calculated by
 \begin{equation}
 \tilde{\Xmat} = \Gmat\Xmat.
 \end{equation}
 $\tilde{\Xmat}$ is an $\left(N - i + 1\right) \times M$ matrix, each row of which is orthonormal to each other and the already estimated components. Letting $\tilde{\wvec}$ be the vector separating the first independent component of $\tilde{\Xmat}$, the vector separating the $i$-th independent component of $\Xmat$ can be calculated as
 $\wvec = \Gmat^{\intercal}\tilde{\wvec}$.
 By estimating $\tilde{\wvec}$ from $\tilde{\Xmat}$ instead of $\Xmat$, the computational complexity is reduced to $\frac{N-i+1}{N}$. In addition, the orthogonalization process is removed from each iteration.
 Furthermore, this process is expected to improve the solution by reducing the search space. The improvement is investigated in Section \ref{results}.
 
\begin{algorithm} 
  \caption{Fast Ordering ICA Using Matrix Representation}\label{alg_algorithm_fast}
    \begin{algorithmic}[1]
    \Require{$\Xmat$: whitened observed signals. $L, K, \epsilon$: hyperparameters.}
    \Ensure{$\Wmat$: separating matrix.}
     \State Initialize $\Wmat$ to the empty matrix
     \State $i \gets 1$
     \While{$i \le N$}
     \If{$i > 1$}
     \State $\Fmat \gets \Imat_{N}-\Wmat^{\intercal}\Wmat$
     \State Delete the bottom $i-1$ rows of $\Fmat$
     \State $\Gmat \gets \left(\Fmat\Fmat^{\intercal}\right)^{-\frac{1}{2}}\Fmat$
     \State $\tilde{\Xmat} \gets \Gmat\Xmat$
     \Else
     \State $\tilde{\Xmat} \gets \Xmat$
     \State $\Gmat \gets \Imat_{N}$
     \EndIf
     \State Initialize $\Bmat$ as an $L \times (N-i+1)$ random matrix
     \State Initialize $\Bmat_{\rm{conv}}$ to the empty matrix
     \State Normalize every $\bvec^{l}$ ($l$-th row of $\Bmat$) by $\bvec^{l}/\sqrt{\bvec^{l}\left(\bvec^{l}\right)^{\intercal}}$
     \State $t \gets 0$
     \Do
     \State $\Bmat_{\rm{prev}} \gets \Bmat$
     \State $\Zmat \gets \Bmat\tilde{\Xmat}$
     \State $\Bmat \gets \left(\Zmat \circ \Zmat
     \circ \Zmat\right)\tilde{\Xmat}^{\intercal}/M - 3\Bmat$
     \ForAll{$l$-th row $\bvec^{l} \in \Bmat$}
     \State Normalize $\bvec^{l}$ by $\bvec^{l}/\sqrt{\bvec^{l}\left(\bvec^{l}\right)^{\intercal}}$
     \State Let $\bvec_{\rm{prev}}^{l}$ be $l$-th row of $\Bmat_{\rm{prev}}$
     \If{$||\bvec^{l} + \bvec_{\rm{prev}}^{l}|| > \epsilon$ or
       $||\bvec^{l} - \bvec_{\rm{prev}}^{l}|| > \epsilon$)}
        \State Remove $\bvec^{l}$ from $\Bmat$
        \State Concatenate $\bvec^{l}$ to $\Bmat_{\rm{conv}}$ as a new row
     \EndIf
    \EndFor
     \State $t \gets t + 1$
     \doWhile{$t < K$ and $\Bmat$ is not empty}
     \State $\Zmat \gets \Bmat_{\rm{conv}}\tilde{\Xmat}$ 
     \State $\alpha_{l} \gets \sum_{m}z_{lm}^{4}/M - 3$ for every $l$ ($z_{lm}$ is each element of $\Zmat$)
    \State $p \gets \mathrm{argmax}_{l}\Upsilon\left(\alpha_{l}\right)$
    \If{$\Upsilon\left(\alpha_{p}\right) < 2\left(N-i+2\right)\left(N-i+1\right)/M$}
    \State \textbf{break}
    \EndIf
    \State Let $\bvec_{\rm{conv}}^{p}$ be $p$-th row of $\Bmat_{\rm{conv}}$
     \State Concatenate $\bvec_{\rm{conv}}^{p}\Gmat$ to $\Wmat$ as the last row
     \State $i \gets i + 1$
     \EndWhile
  \end{algorithmic}
 \end{algorithm}

\subsection{Complete Algorithm}
The complete fast algorithm of the ordering ICA is given in Algorithm \ref{alg_algorithm_fast} using the above three improvements. It does not include any explicit parallel computing process. The computational complexity of Algorithm \ref{alg_algorithm_fast} depends on line 20. If $L$ is sufficiently large and is of the same order as $O\left(N\right)$, efficient matrix multiplication algorithms can be utilized. For example, when the classical Strassen algorithm \cite{Skiena.2008} is used, the complexity is of the order of $O\left(L^{1.8}M\right)$ instead of $O\left(L^{2}M\right)$. Therefore, Algorithm \ref{alg_algorithm_fast} is more efficient than Algorithm \ref{alg_background_original} from a computational theoretical point of view. The practical comparison is shown in Section \ref{results}. 

\section{Results}\label{results}
Here, we show two experimental results: artificial datasets and EEG analysis.

\subsection{Basic Settings}\label{results_settings}
All the experiments were carried out on a 40-core server (dual CPU with 20 cores) with 512GB RAM,
where each core is an Intel Xeon 2.3GHz processor.
All codes were performed with
MATLAB R2023b. 
The proposed method in Algorithm \ref{alg_algorithm_fast} was implemented without Parallel Computing Toolbox. The code is named Fast Ordering.
The implementation of the original ordering
ICA (named Threads Ordering) in Algorithm \ref{alg_background_original} 
uses Parallel Computing Toolbox, which could utilize 40 threads in this environment. 
We also implemented the original ordering ICA by only a single thread (named Single Ordering) without Parallel Computing Toolbox.
Regarding the hyperparameters, 
$\epsilon$ was set to $10^{-6}$.
The maximum number of iterations $K$ was set to 30.
All of the above three codes (with slight modifications) are given at  \url{https://github.com/yoshitatsumatsuda/orderingICA/}.  
In addition, we used the extended InfoMax of EEGLAB 2021.1 (named InfoMax)
at \url{https://github.com/sccn/eeglab} and the Fast ICA using kurtosis (named Fast ICA) at \url{https://research.ics.aalto.fi/ica/fastica/}.
Note again that the algorithms other than Threads Ordering do not need Parallel Computing Toolbox and do not include explicit implementations utilizing multiple threads.
\begin{figure}[htbp]
  \centering
 \begin{tabular}{cc}
  \shortstack{(a) Ordering error\\
  \includegraphics[width=55mm]{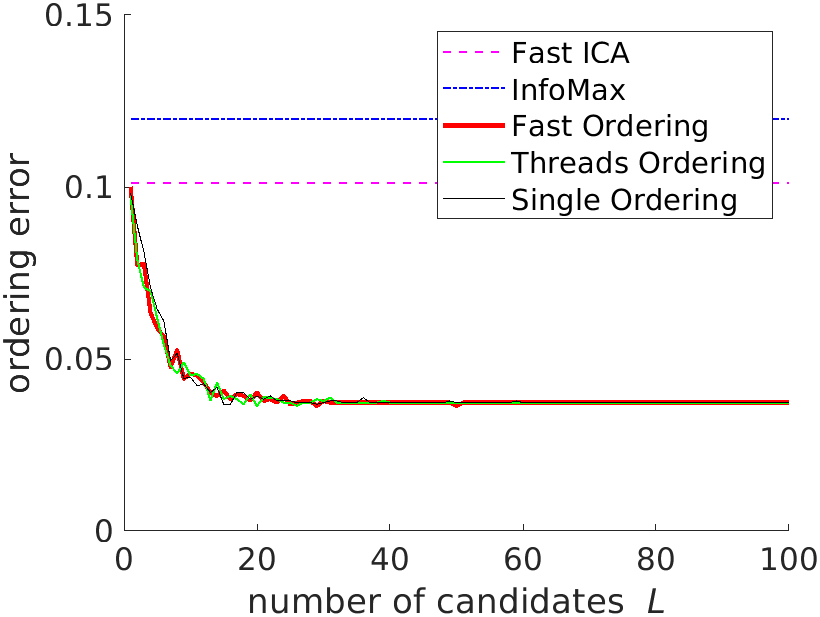}}&
  \shortstack{(b) Calculation time\\
  \includegraphics[width=55mm]{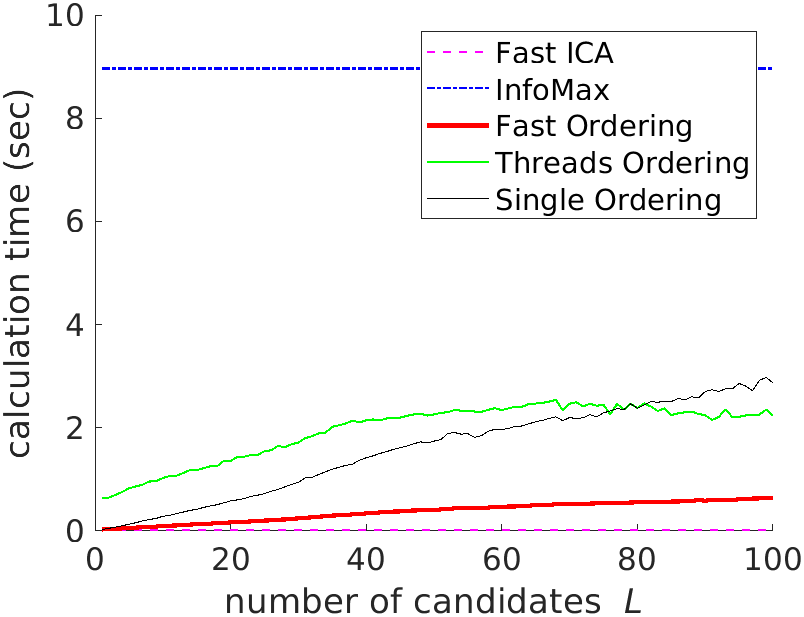}}
  \end{tabular}
 \caption{{\bf Estimation of the order of sources in artificial datasets ($N = 20$).}
 (a) shows the change of $E_{\rm{ordering}}$ of each method along the number of candidates $L$ in the ordering ICA. (b) shows the calculation time of each method along $L$. Regarding InfoMax and Fast ICA, the values form horizontal lines because they do not employ multiple candidates.}
\label{fig_results_artificial_ordering}
\end{figure}
\subsection{Experiments on Artificial Datasets}\label{results_artificial}
Here, generalized Gaussian distributions are used as non-Gaussian sources, whose probability density function is given as
$p(u) =
  \rho\exp\left(-\left(|u|/\beta\right)^\rho\right)/2\beta\Gamma\left(1/\rho\right)$. $\Gamma$ is the
gamma function. The parameter $\beta$ was set so that the variance is 1.
The parameter $\rho$ determines the non-Gaussianity of each source.
$\rho = 2$ corresponds to the Gaussian distribution.
$\rho$ was set to 20 different values
$\rho = 2*2^{i/4}$ for $i = -10, -9, \cdots, -1$ (10 sources with positive kurtoses) and $i =1, 2,  \cdots, 10$ (10 sources with negative kurtoses). 

In the first experiment, we used blind source separation of the above 20 sources with different non-Gaussianity (namely, $N = 20$). The sample size $M$ was set to 10000.
As it is a simple separation problem, many ICA methods can separate every source.
However, only the ordering ICA can estimate the sources in the unique order of non-Gaussianity and does not suffer from the permutation ambiguity problem.
Therefore, we employed $E_{\rm{ordering}} = ||\Wmat\Amat - \Imat_{N}||_{0}/N^{2}$ (named the ordering error), where the sources are assumed to be sorted by the rank of non-Gaussianity.
This error is sensitive to permutation. It is zero only if the order of estimated sources is the same as that of the original sources. 
Fig. \ref{fig_results_artificial_ordering} shows the results of the ordering error and calculation time of the five methods (Fast Ordering, Threads Ordering, Single Ordering, InfoMax, and Fast ICA).
The number of multiple candidates $L$ was set from 1 to 100.
All the values were averaged over 10 runs of random initializations.
Fig. \ref{fig_results_artificial_ordering}-(a) shows that the ordering error converges to a small value by all three ordering ICA methods (Fast Ordering, Threads Ordering, and Single Ordering) if $L$ is more than 20.
On the other hand, InfoMax and Fast ICA could not find the sources in the appropriate order. Fig. \ref{fig_results_artificial_ordering}-(b) shows that Fast Ordering was much faster than Threads Ordering and Single Ordering in this experiment. Although the calculation time is increased according to the number of multiple candidates in Fast Ordering and Single Ordering, the increase of Fast Ordering is much lower than Single Ordering. Even for a small number of candidates, Threads Ordering needed a large overhead time.

\begin{figure}[tbp]
  \centering
 \begin{tabular}{cc}
  \shortstack{(a) Estimated number of non-Gaussian sources\\
  \includegraphics[width=55mm]{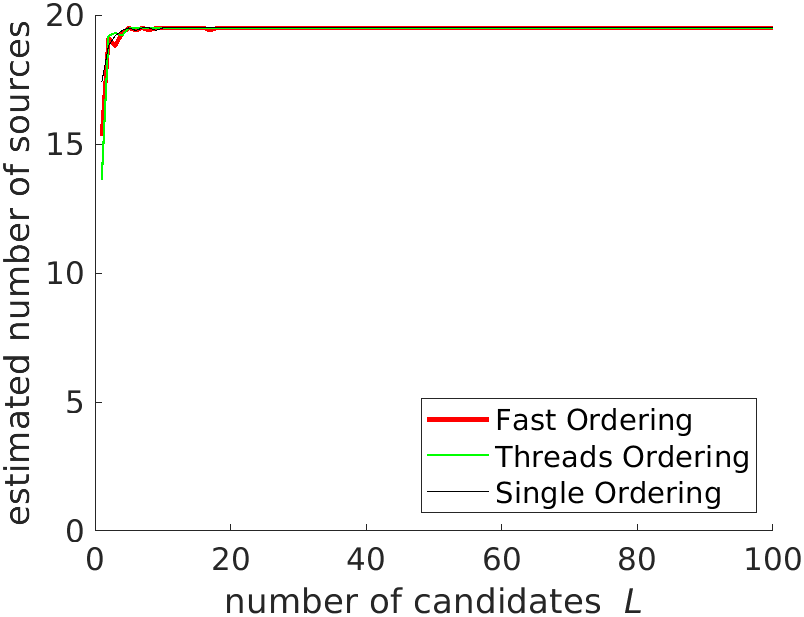}}&
  \shortstack{(b) Calculation time\\
  \includegraphics[width=55mm]{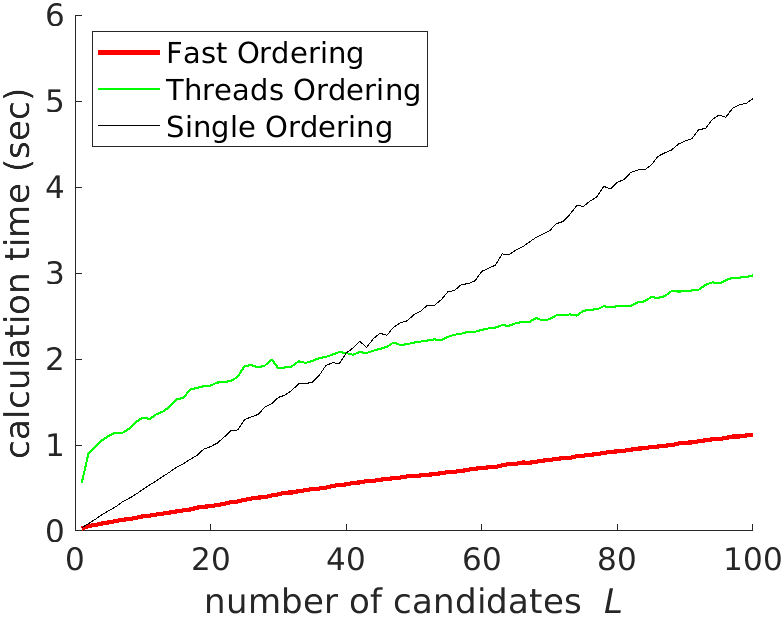}}
  \end{tabular}
 \caption{{\bf Estimation of the number of non-Gaussian sources in artificial datasets ($N = 30$, including 20 non-Gaussian sources and 10 Gaussian noises).}
 (a) shows the estimated number of non-Gaussian sources by the Gaussianity test along the number of candidates $L$ in the ordering ICA. (b) shows the calculation time of each method along $L$.}
\label{fig_results_artificial_number}
\end{figure}

In the second experiment, we employed the Gaussianity test to estimate the number of non-Gaussian sources. 10 Gaussian noises were added to the above 20 non-Gaussian sources so that $N = 30$.
The other settings are the same as those in the first experiment.
Only the three ordering ICA methods were used.
Fig. \ref{fig_results_artificial_number}-(a) shows that the true number of the non-Gaussian sources (10) could be estimated by all three ordering ICA methods if $L$ is more than 20. Fig. \ref{fig_results_artificial_number}-(b) also shows that Fast Ordering was much faster than the other methods.

In summary, the results on the artificial datasets verified that 
Fast Ordering can find almost the same solutions as the previous ordering ICA methods, much more efficiently.
\begin{figure}[tbp]
\centering
\includegraphics[width=\textwidth]{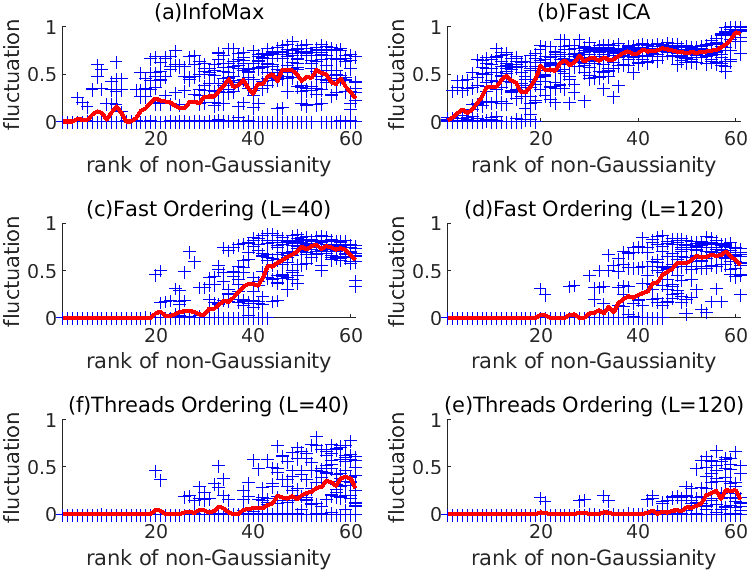}
   \caption{{\bf Fluctuation of solutions of ICA in EEG analysis by InfoMax, Fast ICA, Fast Ordering ($L$ = 40 and 120), and Threads Ordering ($L$ = 40 and 120).} Each point (denoted by '+') corresponds to the value of fluctuation in each EEG signal. The horizontal index is sorted by the rank of the non-Gaussianity of the extracted components.
   The solid curve shows the average of the fluctuations over the 10 EEG signals.}
  \label{fig_fluctuation_5subjects}
\end{figure}
\subsection{EEG Analysis}\label{results_EEG}
EEG analysis is one of the most important practical applications of ICA. 
As the ordering ICA can estimate the source signals uniquely and reliably by avoiding local minima, it is promising in this field \cite{PL1.Matsuda.2022}.
Here, Fast Ordering is applied to EEG data in comparison with Threads Ordering, InfoMax, and Fast ICA. We used the 5SUBJECT dataset at \url{https://sccn.ucsd.edu/eeglab/download/STUDY5subjects.zip}.
It is a widely used benchmark of EEG data, which consists of the 10 EEG signals syn\{02,05,07,08,10\}-s\{253,254\}.
  They were observed in a semantic task with five subjects under two conditions.
  The number of channels is fixed to 61, namely, $N = 61$.
The sample size $M$ of each signal is about 160,000-190,000.
Except for Single Ordering (which is too slow in this problem), the four methods (Fast Ordering, Threads Ordering, InfoMax, and Fast ICA) were applied directly to each EEG signal. As the Gaussianity test was not satisfied in almost all components (in other words, there was no Gaussian noise), it was omitted in this experiment. In addition, the extracted components in InfoMax and Fast ICA were also sorted by the rank of the non-Gaussianity $\Upsilon$.
The number of candidates $L$ in Fast Ordering and Threads Ordering was set in increments of 10, from 10 to 120.

We calculated the average fluctuations of the solutions over the
different runs to evaluate the reliability of the methods.
The fluctuations are measured in the same way as in \cite{PL1.Matsuda.2022}, as explained below.
Each method was applied to each EEG signal from different initializations over $T = 10$ runs. 
Here, $\wvec_{i}$ in the $t$-th run is denoted as $\wvec_{i}^{t}$ ($t = 1, \ldots, 10$).
The divergence $\delta\left(i,t,u\right)$ between $\wvec_{i}^{t}$ and $\wvec_{i}^{u}$ is defined as
$\delta\left(i,t,u\right) = 1 - \left|S_{\cos}\left(\wvec_{i}^{t},\wvec_{i}^{u}\right)\right|$
where $\left|S_{\cos}\left(\wvec_{i}^{t},\wvec_{i}^{u}\right)\right|$ is 
the absolute value of the cosine similarity of $\wvec_{i}^{t}$ and $\wvec_{i}^{u}$.
$\delta\left(i,t,u\right)$ is the minimum 0 if and only if $\wvec_{i}^{t}$ and $\wvec_{i}^{u}$ have the same or opposite directions.
When $\wvec_{i}^{t}$ is orthogonal to $\wvec_{i}^{u}$,
$\delta\left(i,t,u\right)$ is the maximum 1.
Then, the fluctuation of $\wvec_{i}$ over all the runs is evaluated as the average of $\delta\left(i,t,u\right)$ over all the $T(T-1)$ pairs of $t$ and $u$.
\begin{figure}[tbp]
\centering
\includegraphics[width=\textwidth]{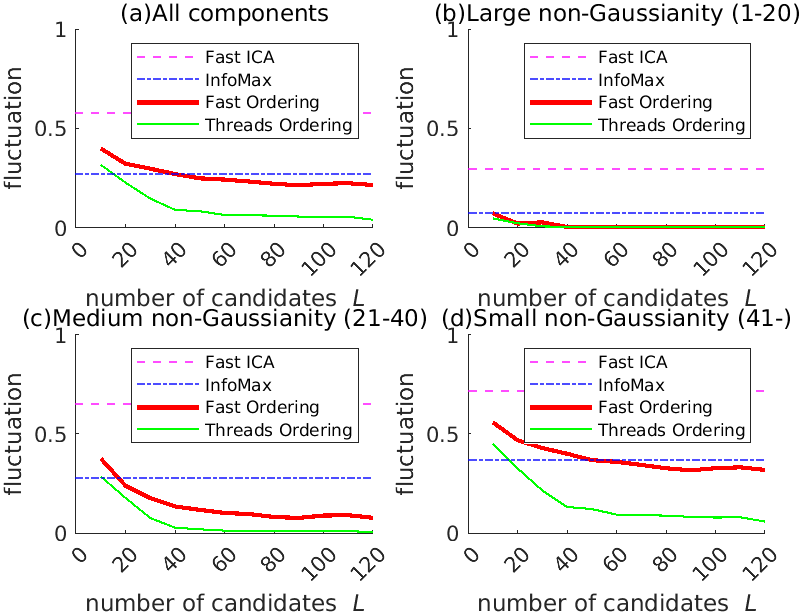}
\caption{{\bf Relation between the number of candidates and the average fluctuation in EEG analysis.} The solid curve shows the average fluctuation over a group of extracted components (all, with large non-Gaussianity (the first-20th ranks), with medium non-Gaussianity (the 21st-40th ranks), with small non-Gaussianity (the rest)). For comparison, the average fluctuations of Fast ICA and InfoMax are displayed by the horizontal lines.}
  \label{fig_details_5subjects}
\end{figure}

First, we examine the fluctuations for the four methods.
Fig. \ref{fig_fluctuation_5subjects} shows the plot of the fluctuations of the 10 signals by InfoMax, Fast ICA, Fast Ordering ($L = 40, 120$), and Threads Ordering ($L = 40, 120$).
Fig. \ref{fig_details_5subjects} shows the details of the relations between the number of candidates $L$ and the average fluctuation.
The fluctuations are averaged over the four different groups of components: all, the top-ranked 20 ones with large non-Gaussianity, the medium-ranked 20 ones, and 
the rest with small non-Gaussianity.
Figs. \ref{fig_fluctuation_5subjects} and \ref{fig_details_5subjects} show that the ordering methods with $L = 40$ can estimate about 20 unique components with high non-Gaussianity. On the other hand, Fast Ordering seems inferior to Thread Ordering in estimating the unique components with relatively low non-Gaussianity concerning the fluctuations.

\begin{figure}[tbp]
  \centering
 \begin{tabular}{cc}
  \shortstack{(a) Degree of non-Gaussianity\\
  \includegraphics[height=43mm]{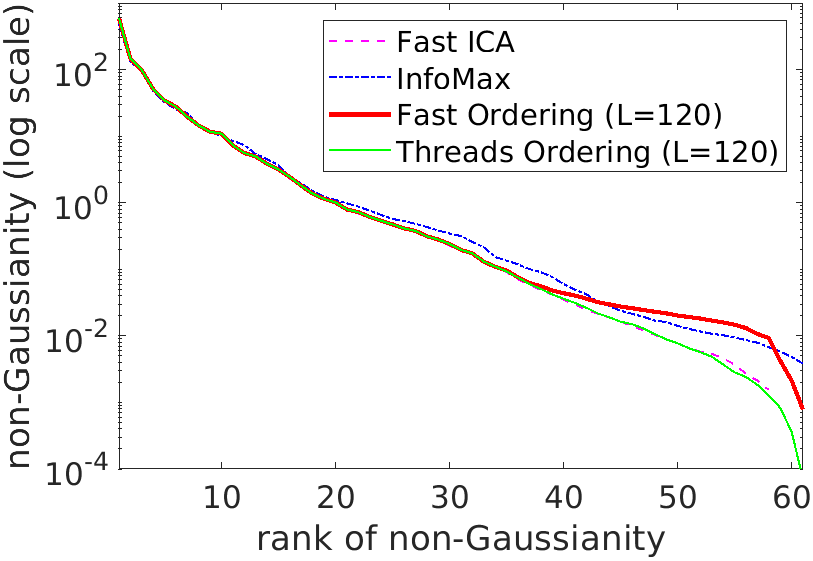}}
\shortstack{(b)  Relative ratio of non-Gaussianity\\
  \includegraphics[height=43mm]{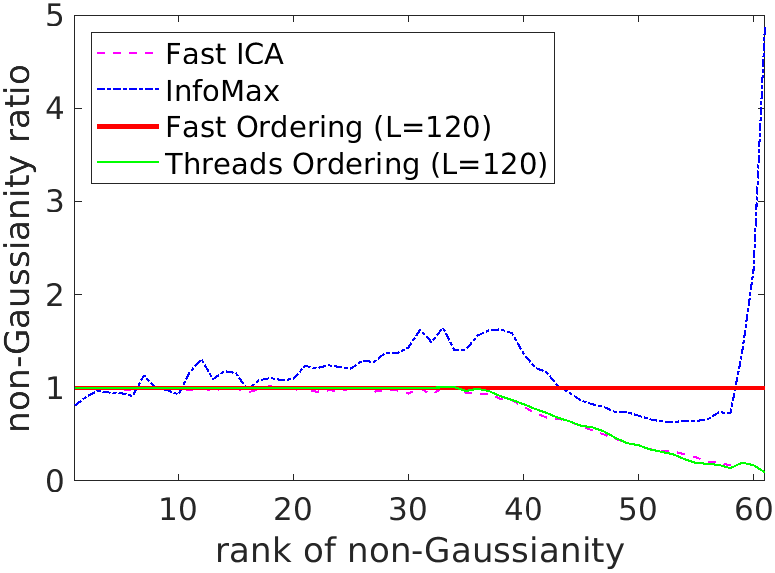}}&
  \end{tabular}
 \caption{{\bf Comparison of non-Gaussianity in EEG analysis for Fast Ordering ($L = 120)$, Threads Ordering ($L = 120$), Fast ICA, and InfoMax.}
 (a) shows the average degrees of the non-Gaussianity $\Upsilon$ (on a log scale) for each sorted component over the 10 EEG signals. (b) shows the relative ratio of the non-Gaussianity to the degrees of Fast Ordering ($L = 120)$.}
\label{fig_non_Gaussianity_5subjects}
\end{figure}

Next, we examine the optimality of the solutions.
Fig. \ref{fig_non_Gaussianity_5subjects} shows the averages of the non-Gaussianity $\Upsilon$ of the extracted components for Fast Ordering with $L = 120$, Threads Ordering with $L = 120$, InfoMax, and Fast ICA.
The components are sorted by the non-Gaussianity in each signal. 
 Fig. \ref{fig_non_Gaussianity_5subjects} shows that Fast Ordering could find the same or more optimal solutions than Threads Ordering. Because InfoMax often finds relatively worse solutions for the components with high non-Gaussianity, it is natural that the solution of InfoMax is often better than the ordering ICA methods for the components with relatively low non-Gaussianity. 
 On the other hand, the solution of Fast Ordering is always the same or better than Threads Ordering. This is probably because the search space for Fast Ordering is appropriately reduced, as explained in Section \ref{algorithm_reduction}. 
 The improved maximization of the non-Gaussianity and the increase of the fluctuations in Fast Ordering suggest that a stable global optimum does not exist if the non-Gaussianity of a component is quite low. This property may be useful for extracting only the essential components from EEG signals without the Gaussianity test.

\begin{figure}[tpb]
\centering
\includegraphics[width=\textwidth]{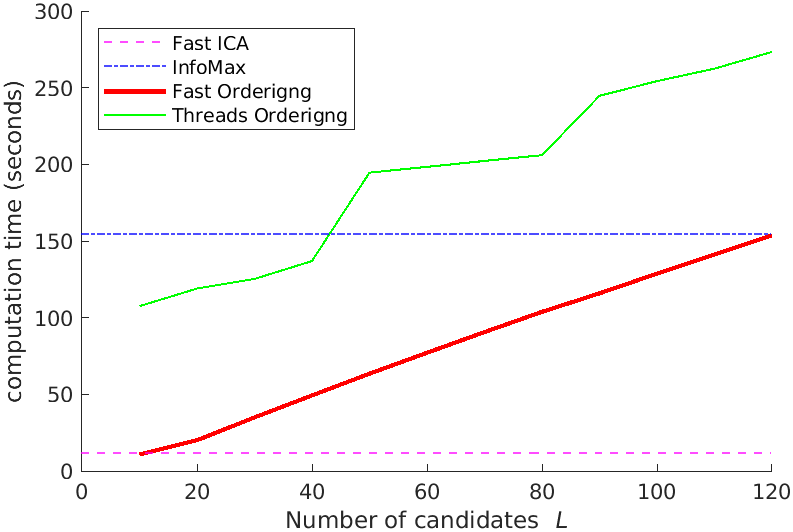}
   \caption{{\bf Calculation time in EEG analysis.} The solid curve shows the average calculation time over all the signals by the number of candidates $L$ for Fast Ordering and Threads Ordering.
   For comparison, the horizontal lines show the average calculation time of Fast ICA and InfoMax.}
  \label{fig_time_5subjects}
\end{figure}

Finally, we examine the calculation time.
Fig. \ref{fig_time_5subjects} shows the relation between the calculation time of the two ordering ICA methods and the number of candidates $L$ compared to Fast ICA and InfoMax.
Fig. \ref{fig_time_5subjects} shows that Fast Ordering was faster than InfoMax and Threads Ordering if $L$ is less than 120.
Fast Ordering was about three times as fast as InfoMax and Threads Ordering for $L = 40$.
In other words, Fast Ordering is much superior to the other methods in extracting the top-ranked 20 non-Gaussian components from EEG signals.

\section{Conclusion}\label{conclusion}
In this paper, we proposed a novel algorithm for implementing the ordering ICA by matrix representation. Experimental results verified that the proposed algorithm is much faster than the previous ones. Moreover, the proposed algorithm could find more optimal solutions in EEG analysis even if the non-Gaussianity of the component is quite low.
This phenomenon is expected to lead to the extraction of only the essential components from EEG data. We are planning to investigate this phenomenon further. We are also planning to utilize the ordering ICA in many practical applications. Moreover, we are now developing a method to increase the robustness of the kurtosis-based ICA by removing a few extreme samples. This work was supported by JSPS KAKENHI Grant Number JP24K15093.

\bibliographystyle{splncs04}
\bibliography{reference}
\end{document}